\documentclass[runningheads]{llncs}
\usepackage{graphicx}
\usepackage{amsmath}
\usepackage{multirow}	
\usepackage{xcolor, colortbl}
\definecolor{lg}{gray}{0.9}

\begin{document}
\title{Detecting sexually explicit content in the context of the child sexual abuse materials (CSAM): end-to-end classifiers and region-based networks}
\titlerunning{Detecting sexually explicit content in the context of CSAM}

\author{Weronika Gutfeter\inst{1}\orcidID{0000-0001-6359-8220} \and
Joanna Gajewska\inst{1}\orcidID{0009-0004-8760-5682} \and
Andrzej Pacut\inst{1}\orcidID{0000-0003-3489-8990}}

\authorrunning{W.Gutfeter et al.}

\institute{Research and Academic Computer Network (NASK) \\
Kolska 12, 01-045 Warsaw, Poland
\url{www.nask.pl} }
\maketitle              
\begin{abstract}

Child sexual abuse materials (CSAM) pose a significant threat to the safety and well-being of children worldwide. Detecting and preventing the distribution of such materials is a critical task for law enforcement agencies and technology companies. As content moderation is often manual,  developing an automated detection system can help reduce human reviewers' exposure to potentially harmful images and accelerate the process of counteracting. 
This study presents methods for classifying sexually explicit content, which plays a crucial role in the automated CSAM detection system. Several approaches are explored to solve the task: an end-to-end classifier, a classifier with person detection and a private body parts detector. All proposed methods are tested on the images obtained from the online tool for reporting illicit content. Due to legal constraints, access to the data is limited, and all algorithms are executed remotely on the isolated server. The end-to-end classifier yields the most promising results, with an accuracy of 90.17\%, after augmenting the training set with the additional neutral samples and adult pornography. 
While detection-based methods may not achieve higher accuracy rates and cannot serve as a final classifier on their own, their inclusion in the system can be beneficial. Human body-oriented approaches generate results that are easier to interpret, and obtaining more interpretable results is essential when analyzing models that are trained without direct access to data.

\keywords{CSAM  \and NSFW \and explicit content \and child pornography \and pornography  \and body part detection.}
\end{abstract}

\section{Introduction}
Methods for classifying sexually explicit content in images proposed in this paper are part of a larger project whose goal is to detect child sexual abuse materials (CSAM). Child sexual abuse materials are often incorrectly referred to as child pornography. Field experts usually prefer the term CSAM as it puts weight on the harm that is inflicted on minors when generating or sharing such types of images.
There are several challenges that need to be solved to build a robust CSAM detector. The most significant factor is limited access to the training and evaluation data, which is illegal to possess and can be obtained only by authorized governmental or non-governmental agencies. Some of the concepts can be transferred from models trained on adult pornography, but we believe that this approach does not give a full perspective on the topic. Secondly, pornography itself is difficult to be strictly defined and the definitions may vary across countries. Differences between images that contain sexually explicit content and other socially acceptable images can be very subtle and context-dependent.

In this study, we start with the analysis of data that is collected and annotated by the experts working with illegal materials on a daily basis in the public agency named NASK-Dyżurnet. Some of the main concepts of our work are data-driven. First, we analyzed the content using its annotations and consequently, we proposed a classification schema to be implemented in the CSAM detection system. Experiments with classification are conducted for the three considered approaches. The first approach is an end-to-end classifier that is built with a single neural network. In the second approach, we classify patches containing human silhouettes found by the person detector. Then, we discuss the solution for applying the private body part detector. In the last part, we discuss all the methods and show some of the qualitative results. Because of the delicate nature of data, we can present only false positive results for neutral images or adult pornography.

\section{CSAM detection methods} 

There are several approaches to CSAM and pornography detection proposed in the literature and among publicly available libraries and services. One of the common methods is to build a database of files that are already marked as illegal. A database can be indexed with the metadata or hashes of the images like in the Microsoft PhotoDNA~\cite{photodna}. Content-based blacklists are a powerful tool, but they can be vulnerable to some set of heavier image modifications and cannot be used to recognize newly generated data. For that reason, many methods that make use of machine learning models were proposed. Some of the solutions are based on nudity or skin area detection~\cite{nudetective},\cite{towardchildporn}. However, it is necessary to emphasize that, in our case, nudity is not sufficient to label the image as illegal. A more important factor, but also not the only one, is the visibility of private body parts in the image, like genitalia and anal area. Private body parts detectors are frequently used in the systems for pornography detection described in the literature~\cite{ppcensor},\cite{nudenet},\cite{sod}. Authors of the model proposed in~\cite{ppcensor} train their detector to find in the image male or female genitalia, buttocks, and breasts. In~\cite{sod}, authors additionally distinguish whether the genitalia are involved in the sexual activity or exposed without sexual context.

Another typical approach to identifying pornographic or harmful content is to build a classification model that assigns the input into one of the selected categories. Recently, classification models are most commonly deep neural networks or even ensembles of networks~\cite{attm-cnn},\cite{leveragingdnn2fight},\cite{nudenet},\cite{ppcensor},\cite{thorn}. Among solutions, we can distinguish end-to-end neural classifiers that take the image as an input and return its label.  Models can be either binary classifiers that answer the question of whether the image belongs to CSAM or can identify the other classes important for authors, like adult pornography or erotic content. Definitions of categories can vary. In model from~\cite{nudenet} content is classified as safe or unsafe to work with (NSFW), in \cite{thorn} into CSAM, adult pornography and neutral images, in the model from~\cite{nudetective} into images containing nudity and neutral ones. End-to-end models trained on representative data can gain high-accuracy results, but their decisions can be problematic to interpret. 

A solution to that problem is to decompose the classification task into several subtasks. The authors of~\cite{attm-cnn} proposed the model that detects CSAM using two submodels: one detects children and the second looks for pornographic content. The final classification can be obtained by merging the information from the two submodules.  A very similar approach is presented by the authors of~\cite{leveragingdnn2fight}. Their proposed model has two stages. In the first stage, they use a neural network to decide whether an image is pornographic or not, and in the second stage, they apply SVM to distinguish between adult pornography and CSAM. 

As the multi-module approaches are more elastic than the single-one, we also decided to separate the task of detecting children from identifying the sexual context of the image.

\section{Source of illegal images}

Our work is a result of collaboration with the public agency named NASK-Dyżurnet~\cite{dyzurnet}, dedicated to fighting the crime of sharing CSAM in cyberspace. Collaboration is a part of a project to build an automatic system for CSAM detection. Experts from the NASK-Dyżurnet collect data of various types, including images, video, and text, and classify them into legal or illegal content.  They proposed an annotation system and applied a selected set of labels to the collected data. The source of data is mainly reports from internet users. 

The annotating procedure has two stages: the first is coarser and consists of labeling images with one of the global CSAM classes. Global CSAM classes belong to the following list of categories: adult pornography, child erotism, child nudity, sexual posing (CSAM), minors only (CSAM), minors~\&~adults (CSAM), in the presence of a minor (CSAM), focus (CSAM), other (CSAM) and other neutral (not-CSAM). There is also a set of additional tags for the image that can refer to the type of sexual activity or the age of the youngest person, but without indicating its source.  In the second stage, more detailed information is added, like bounding boxes of persons or body parts visible in the image. Bounding boxes also have additional attributes like the person's age, sex, and visible activity. Body parts are one of the following: female genitalia, male genitalia, anal area, or breasts.  However, in our experiments, breasts are ignored as their presence is not included in our definitions of sexually explicit content. Creating second-stage annotation is more time-consuming, and therefore, only a subset of the whole dataset collected by the NASK-Dyżurnet is annotated. We will refer to these two parts as DN-A and DN-C. 
DN-A is a part of data that contains a full set of annotations from the first and the second stages. It has both image-related annotations and bounding boxes of persons and body parts. The remaining part of the data will be named DN-C and it is a part with image-related annotations, but without bounding boxes. DN-C is considered as weakly annotated because not for every image we can conclude whether it contains sexual activity or not using a given set of tags. Results discussed in this paper are evaluated only on the DN-A.

Access to the DN data is restricted because of legal issues, and only people who work inside NASK-Dyżurnet teams are allowed direct insight. The authors of this paper can run the models on the isolated server and analyze the aggregated results. It is not possible to watch the restricted images and the data itself is mostly illegal. Legal images can be previewed. Not-CSAM examples from the DN dataset can still be described as hard because, in most cases, they contain naked adults or children or soft erotics. Moreover, the number of samples in image categories proposed by NASK-Dyżurnet experts is not equally distributed. For example, a set labeled as "in the presence of a minor (CSAM)" has less than 50 samples, and "child erotism"  has about 250 samples. Bias is also observed for the age and sex of persons visible in images. About 75\% of the bounding boxes with persons are female, 45\% are female minors, and only 3\% are male minors. Due to the imbalance, we propose our own classification schema. Additionally, information about age and sex is used when drawing folds for cross-validation tests to ensure that samples from each group can be found in each evaluation set.

\begin{table}
\caption{Contents of the dataset DN used to evaluate and train the proposed models. The dataset is divided into two subsets: DN-A (with a full set of annotations) and DN-C (only with coarse labels). Neutral samples in the dataset can be considered challenging - we show in this table how many samples are warning neutral in the sense of containing nudity or soft erotics.}
\label{tab:dn-data}
\begin{tabular}{|p{0.3\textwidth}|p{0.15\textwidth}<{\raggedleft}|p{0.15\textwidth}<{\raggedleft}|p{0.15\textwidth}<{\raggedleft}|p{0.2\textwidth}<{\raggedleft}|}
\hline
\multirow{2}{*}{Dataset} & \multicolumn{4}{c|}{N samples} \\ \cline{2-5}
 & All & Sexual activity & Sexual posing & Neutral (warning)\\ \hline
DN-A & 7688 & 2203 & 2715 & 2770 (1206) \\
DN-C & 11578 & 972* & 4152* & 6454 (1887) \\
Sum & 19266 & & & \\\hline
\multicolumn{5}{p{\textwidth}}{*DN-C should be treated as a weakly annotated dataset - some of the images marked here as sexual posing can contain sexual activity, which is not reflected in the annotations.  Neutral labels are exact. DN-C dataset is used only in the pretraining phase. } \\
\end{tabular}
\end{table}

\section{Proposed CSAM classification schema}

\begin{figure}[!htb]
\centering
\includegraphics[width=0.8\linewidth]{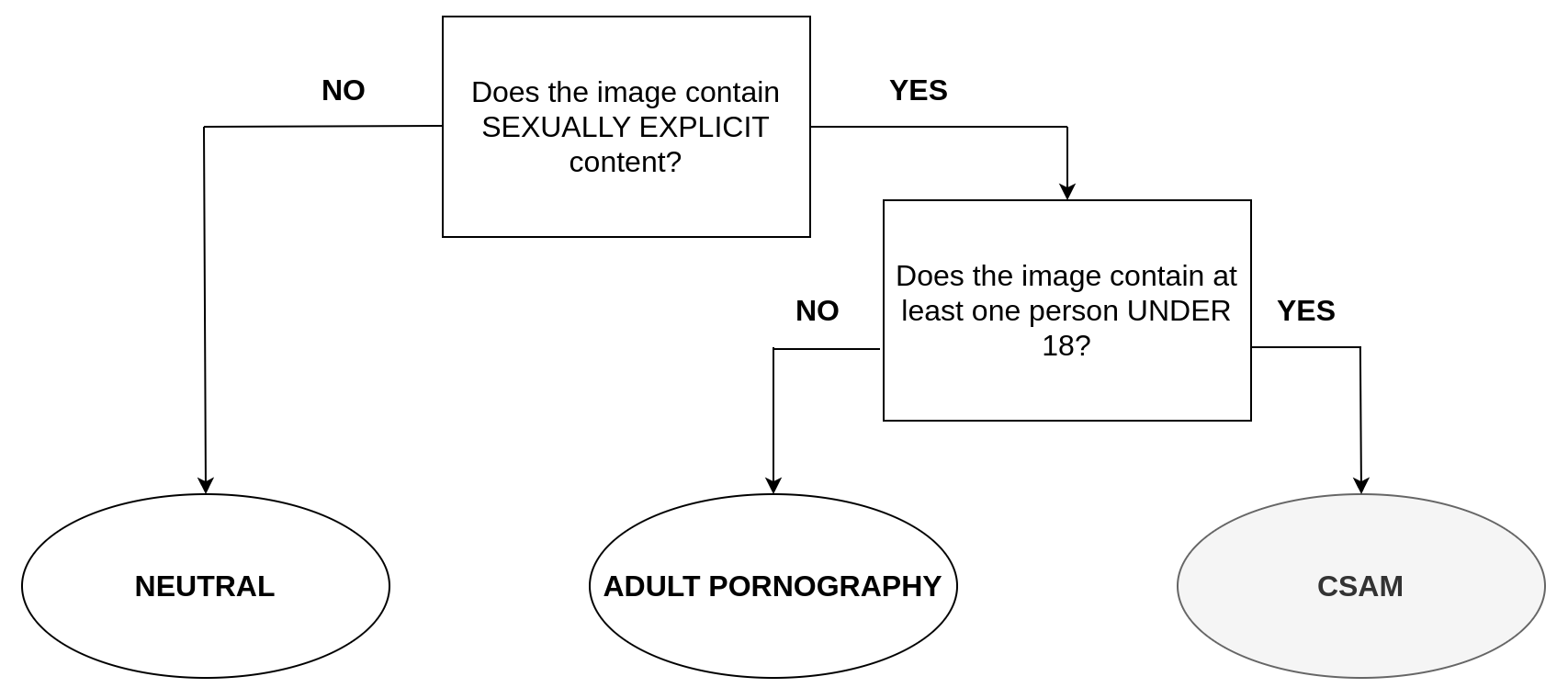}
\caption{Decision schema that leads to the final classification in the proposed CSAM detection system. A complete system consists of two models: one performs age estimation and the second predicts whether the image contains sexually explicit elements (SE classification). Two-stage classification enables the system to distinguish between CSAM, adult pornography and neutral images.}
\label{fig:schema-csam}
\end{figure}

The aim of the complete system is to distinguish between two classes of images: CSAM or not-CSAM, which is equivalent to the decision of whether the image is illegal or not. As was discussed in the former sections, this complex task can be decomposed into two subtasks: one of which answers the question about the presence of minors in images, and the second, which warns about pornographic features. The first subtask can be solved by applying models for age estimation, which is a topic widely discussed in the literature~\cite{lehmann},\cite{age1},\cite{age2},\cite{age3}.  Age can be estimated from the person's face or full silhouette appearance, or other body elements. We won't go deeper into the details of age estimation methods in this paper as we are focusing on the detection of sexually explicit content. Having a classifier that can distinguish between sexually-explicit images (SE) and not-sexual images (NS) together with an age estimator allows us to build a fully functional CSAM detection system. The concept of the two-stage classification procedure is shown in Fig.~\ref{fig:schema-csam}.

The distinction between SE and NS content, named here, SE classification, for brevity, is not a task that is easy to define. A lot of concepts presented in this paper are learned from data, in the sense that we set the classification system after analyzing the data that have been annotated by the experts. We also wanted to be possibly coherent with other sources of pornographic images, which usually don't have such granular classification as data from DN. For example, in dataset Pornography-2K described in ~\cite{braz2k} categories are pornographic or not-pornographic, in paper~\cite{nudenet} images can be safe or unsafe to work with. Therefore, we decided to propose a new dictionary of labels. An additional reason is that the number of samples in some of the native DN categories is small and it would be hard to train the classifier to recognize them. 

Our analysis leads to discriminating two sexually explicit classes, namely sexual activity and sexual posing. The sexual activity category is interpreted as images showing any type of intercourse or interaction that involves intimate body parts of the person or persons visible in the image. Images that belong to the sexual posing class, in most cases, contain only one person who is nude and posing to the camera in a sexual way. Photographs showing sexual posing can be hard to distinguish from neutral nudity. Here, the visibility of the private body parts and suggestive posing is a necessary condition to be labeled as the first category. After distinguishing these two SE classes, we obtained a more balanced dataset for training and evaluation. 

\begin{figure}[!htb]
\centering
\includegraphics[width=0.8\linewidth]{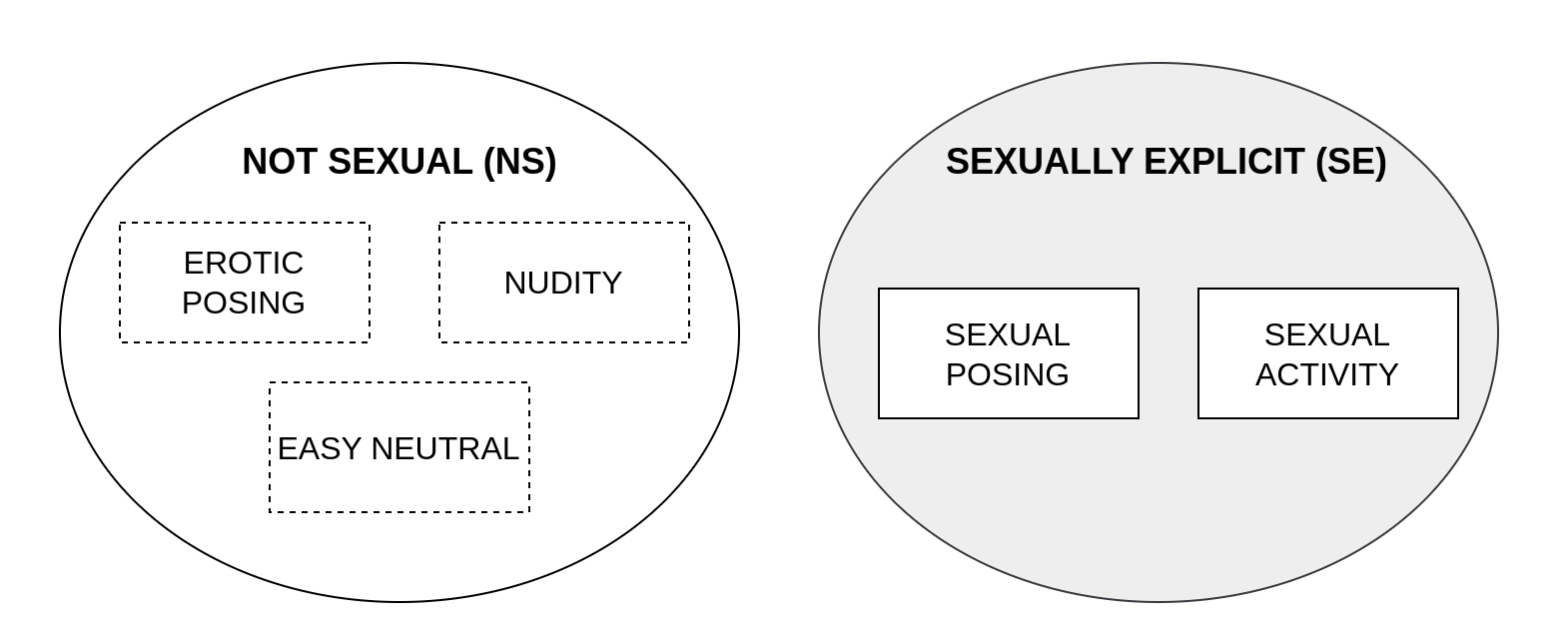}
\caption{Distinguishing between images that contain sexually explicit content (SE) and not-sexual ones (NS) is sufficient for the system to detect CSAM. However, we decided to divide the SE category into images showing sexual activity or sexual posing. The not-sexual group can contain hard samples as images showing neutral nudity and soft erotics.}
\label{fig:schema-sex}
\end{figure}

To sum up, we classify images into 3 categories: sexual activity, which is SE class, sexual posing, which is also SE class, and neutral images, which are NS. The number of samples that belong to each of the categories is given in Tab.:~\ref{tab:dn-data}.  It can be seen that sexual activity is noted in about 29\% of images from the DN-A dataset and sexual posing in 35\%. In the neutral part of the DN dataset, there are a lot of gray area images that, by some terms, can be classified as pornographic. Among neutrals, images showing nudity can appear without erotic context and erotic pictures but without visible nudity. The estimated number of the hard samples is also shown in Tab.~\ref{tab:dn-data}. It can be seen that for the DN-A dataset, almost half of the neutral samples can contain some disputable content. The reason for the high number of hard neutral samples is that data is mostly collected from online reports. The types of content with respect to the proposed classification schema are depicted in Fig.~\ref{fig:schema-sex}. 

Our system can be easily mapped to the COPINE scale proposed in~\cite{copine}. L6 - "Explicit erotic posing" is a sibling class to our "Sexual posing" and L7 - "Explicit Sexual Activity" to "Sexual Activity". All categories below L6 in COPINE would be classified as neutral.

\section{Sexually-explicit content classification: model architectures and training}

Our analysis of SE classification starts from the end-to-end classification models. By the end-to-end solution, we mean a deep neural network with the last layer of the length equal to the number of the proposed categories, in our case, 3. The end-to-end network takes the 3-channel RGB image as its input and generates the most probable category label. We consider several convolutional network architectures for this task.  Two versions of ResNet networks proposed in the paper from 2015~\cite{resnet} were trained: one with 50 and one with 101 layers (ResNet-50 and ResNet-101).  For the comparison, we take two newer models from 2021, EfficientNetV2-S and EfficientNetV2-M~\cite{efficientnet2}. All models were initialized with weights transferred from the model trained on ImageNet~\cite{imagenet}. 

Inputs are classified into three categories, knowing that the binary classification would be sufficient for the whole CSAM detection system. Defining a larger category dictionary than necessary is not a standard approach, therefore we use a hierarchical cross-entropy $ \mathcal{L}_H$ that put weight also on binary classification. Hierarchical cross-entropy, in our case, is a two-component loss function. The first component $ \mathcal{L}_{SE}$ enforces correct classification into 2 categories: SE (sexually explicit) and NS (not-sexual). The second one $\mathcal{L}_3$ controls classification into 3 defined categories: sexual activity, sexual posing and neutral. Parameter $\alpha$ was set to 0.5 after preliminary experiments.

\begin{align}
\begin{split}
\label{eq:hce}
\mathcal{L}_H= & \alpha\mathcal{L}_{SE}+(1-\alpha)\mathcal{L}_3\\
 & =\alpha \sum_{x'}p(x')log(p(x')) + (1-\alpha)\sum_{x^{\prime\prime}}p(x^{\prime\prime})log(p(x^{\prime\prime})), \\
& \text{where }x'\in\{SE,NS\}\\
&\text{and } x^{\prime\prime}\in\{\text{"sexual activity", "sexual posing", "neutral"}\}
\end{split}
\end{align}

 For the evaluation of the models, we use the DN-A dataset. As we do not have direct access to the data and cannot inspect it more carefully, we assume that some of the labels may be incorrect. To diminish the uncertainty, tests are performed with 10-fold cross-validation. Each time, data is divided into two parts for training and validations, with about 20\% of samples taken for validation. Folds are drawn with respect to the source DN categories tagged by NASK-Dyżurnet experts, sex, and age of persons visible in the image. It is made to have samples from all source categories for male and female subjects in each evaluation subset.  An important step taken when curating the dataset was also to exclude images that are near duplicates from the data. For this task, we used the preliminary version of our classification model and identify the images that are close in the sense of Euclidean distance between feature vectors. We observed that they were often the same people posing in different sets of clothes or images cut from the sequence (collage or video).  
 
 We show the model comparison results in Tab.~\ref{tab:results1}. By SE accuracy, we mean the number of samples that are correctly predicted as SE or NS (binary classification), but we also show the number of correctly predicted samples (TPR) in each of the groups from the 3-categories set. After the experiments, we decided to continue our research with model SEN-EM based on architecture EfficientNetV2-M as we did not assume any limitations to the model size. However, it is worth noting that the smaller model SEN-ES version also gave high-accuracy results.

\begin{table}
\caption{Results for experiments with different model architectures. Evaluation is made on dataset DN-A. A selected metric is binary accuracy: sexually-explicit (SE) vs. not-sexual (NS).}
\label{tab:results1}
\begin{tabular}{|p{0.15\textwidth}|p{0.22\textwidth}|p{0.11\textwidth}<{\raggedleft}|p{0.11\textwidth}<{\raggedleft}|p{0.11\textwidth}<{\raggedleft}|p{0.11\textwidth}<{\raggedleft}|p{0.12\textwidth}<{\raggedleft}|}
\hline
\multirow{2}{*}{Model} & \multirow{2}{*}{Architecture} & \multirow{2}{0.11\textwidth}{Accuracy} & \multirow{2}{0.11\textwidth}{F1-score} & \multicolumn{2}{c|}{TPR} &  \multirow{2}{*}{Size}\\ \cline{5-6}
 & & & & Sexual activity & Sexual posing &  \\ \hline
SEN-R50 & ResNet-50 & 86.96\% & 0.894 & 67.88\% & 69.82\% &  97.8 MB \\ 	
SEN-R101 & ResNet-101 & 88.31\% & 0.905  & 70.27\% & 73.96\%	&  170.5 MB \\ 
SEN-ES & EfficientNetV2-S &  89.10\% & 0.912 & 70.52\% & 78.21\% &  82.7 MB \\ 
SEN-EM & EfficientNetV2-M & \textbf{89.57\%} & \textbf{0.916} & \textbf{74.09\%} & \textbf{78.75\%} &  208.0 MB \\ \hline 	

\end{tabular}
\end{table}

\section{Pretraining model with external data}

Working on data with limited access has multiple drawbacks. One of them is problematic analysis of results, and scientists cannot look into the images to find the reason for potential errors. Another problem is that the dataset cannot be easily curated. Bias in data is mainly estimated by the analysis of tags. Preliminary experiments on the DN-A showed that the training set should be extended with neutral image samples. In Fig~\ref{fig:cocofp}, we show samples from the COCO~\cite{coco2015} that were falsely labeled as positive. The system tends to make errors in images depicting people with food or other objects near their mouths or in the presence of a bed. Some images focusing on hands and fingers also appeared among incorrect results. 

Enlarging the number of samples in the training set by adding some external data is a typical step that can be made to leverage the results. In our experiments, we use images from the dataset COCO~\cite{coco2015} as it contains natural images showing people. Moreover, it has annotations of persons that can be used when developing solutions based on region proposal networks. The COCO dataset contains over 120.000 images with 91 category labels, where the category with id 1 stands for the bounding box of a person.
Finding sources of neutral data is relatively easier than collecting SE samples. One of the ways to collect external samples for SE categories is to add images with adult pornography. We used the publicly available dataset  Pornography-2K~\cite{braz2k}, which is a set of video materials cut into frames. Pornography-2K is an extension of a database introduced in~\cite{braz1} containing neutral samples. Neutral samples marked as "not pornographic difficult" show people on a beach, young children and contact sports. The main problem with the not neutral samples in Pornography-2K is that a large number of pornographic frames are not sexually-explicit by our definition. Therefore we needed to use additional annotations created for the purpose of body-part detection by authors of~\cite{ppcensor} to select the frames that have bounding boxes either of genitalia or anal area. As a result, we obtained a set of images that are pornographic in our terms. 
Another dataset that can be used in the pretraining phase is the DN-C, the weakly annotated part of the DN dataset. Because of the partial set of labels, it was not involved in the main phase of training.

After collecting additional data samples, we performed a two-stage training procedure. In the first stage, we use external data from one of the mentioned datasets: COCO, Pornography-2K or DN-C. In the second stage, models are fine-tuned on the images from DN-A. Adding the pretraining phase helps improve the results. In the Tab.~\ref{tab:results-pre}, we show accuracy computed on the DN-A and for comparison also on the COCO and Pornography-2K test sets. For the two of the setups presented in the table, SEN-EM-A and SEN-EM-AB, we got a higher accuracy on DN-A than for the baseline method SEN-EM without pretraining. The best average accuracy is for the model SEN-EM-AB. Surprisingly, adding pornography materials from the Pornography-2K  dataset (model SEN-EM-B) makes the accuracy of the DN-A lower. It may be a sign that datasets with adult pornography differ substantially from the data collected on the NASK-Dyżurnet server. Another reason for the lower quality could be that Pornography-2K contains many redundant frames for training as they come from the video footage. Some further experiments show that if we train the model exclusively on Pornography-2K and COCO, the accuracy on DN-A drops from 89.57\% to 70.99\%. Meanwhile, the accuracy on the Pornography-2K test set changes only a little from 74.47\% to 79.41\%.  DN-A dataset seems to be a more representative source of information for learning the concept of SE classification.

\begin{table}
\caption{End-to-end SE classifiers with additional pretraining phase on external image databases. Grey cells are values that surpassed the results for the baseline model SEN-EM without pretraining}\label{tab:results-pre}
\begin{tabular}{|p{0.21\textwidth}|p{0.32\textwidth}|p{0.11\textwidth}<{\raggedleft}|p{0.1\textwidth}<{\raggedleft}|p{0.1\textwidth}<{\raggedleft}|p{0.1\textwidth}<{\raggedleft}|}
\hline
\multirow{2}{*}{Model} &  \multirow{2}{*}{Pretraining on} & \multicolumn{4}{c|}{Accuracy}\\ \cline{3-6}
 & & DN-A &  COCO &  Porn2K & Avg \\ \hline
SEN-EM & No pretraining &  89.57\% &  98.40\% & 74.47\%  & 87.48\% \\ 
SEN-EM-A & DN-C~+~COCO & \cellcolor{lg}\textbf{90.18\%} &  \cellcolor{lg}99.18\% & \cellcolor{lg}75.89\% &  \cellcolor{lg}88.42\%  \\
SEN-EM-B & Porn-2K~+~COCO & 87.45\% &  \cellcolor{lg}\textbf{99.30\%} &  \cellcolor{lg}78.30\% & \cellcolor{lg}88.35\% \\
SEN-EM-AB & DN-C~+~Porn-2K~+~COCO & \cellcolor{lg}90.17\% & \cellcolor{lg}99.20\%  &\cellcolor{lg}\textbf{81.07\%} &  \cellcolor{lg}\textbf{90.14\%}
 \\ \hline
\end{tabular}
\end{table}

\section{Person detection}

Intuitively, the positions of features that are important for SE classification should be correlated with the positions of the human bodies in the image. Therefore, the classification can also be made on patches that show people. In the patch-based scenario, we can additionally obtain information on whose activities are a reason to label the sample in a certain way. It can be stated whether a single person is naked, posing, or involved in intercourse. In the future, we will consider extending the number of person-related categories and diversifying them from the global set of labels.

In this study, we propose an approach to the classification in which the person detector is employed as the alternative to the end-to-end classifier. The proposed method consists of two stages. Firstly, all persons visible in the image are detected, and then the classifier is applied to each detected image patch. In the end, information is aggregated from all the image patches. To compile the final decision, we calculate the maximum from the values of labels. SE labels are rated by their subjective gravity: neutral images(1), sexual posing(2), and sexual activity(3). In this configuration, if at least one person is described as sexually-explicit (SE), then the whole image is SE. 

Person detection is a task that can be solved by the region-proposal networks like Faster-RCNN\cite{faster-rcnn}, SSD\cite{ssd} or YOLO\cite{yolo1},\cite{yolox}. We tested some off-the-shelf methods of person detection and compared them with the ground-truth bounding boxes from DN-A. We used models from libraries like PyTorch~\cite{pytorch} (SSD and Faster-RCNN with ResNet-50 backbone) and mmdetection~\cite{mmdetection} (YoloX-L). From our observations of publicly available pornographic images, such data can vary from other sources of imaging used for training detection algorithms, and these differences can be challenging. Particularly, images that show intercourse or other sexual activity usually do not contain full body silhouettes, and bodies are occluding themselves. Persons in images labeled as sexual posing are often easier to recognize because they're alone and well-visible. 
Detection results for the three selected algorithms are presented in  Tab.~\ref{tab:persondetectors}. For the evaluation, we use metrics proposed by authors of COCO benchmark~\cite{coco2015} with the IoU (intersection over union) set to 0.5. According to the previous evaluation scenarios, training was repeated 10 times on the same cross-validation folds. The highest average precision (AP) is obtained for the YoloX-L detector and it is used in the following experiments. 
\begin{table}
\caption{Person detection accuracy on dataset DN-A for the selected detection methods: SSD, FasterRCNN (ResNet-50), and YoloX-L. Models are evaluated using average precision for IOU=0.5 and average recall as defined in the COCO benchmark. }\label{tab:persondetectors}
\begin{tabular}{|p{0.36\textwidth}| p{0.2\textwidth}<{\raggedleft}|p{0.2\textwidth}<{\raggedleft}|p{0.2\textwidth}<{\raggedleft}|}
\hline
Model &  AP$^{IOU=0.5}$ &  AR & Size\\ \hline
SSD & 0.819 &  0.941 &  136.0 MB \\
FasterRCNN & 0.843 &  0.947 &  167.1 MB \\
YoloX-L & 0.845 &  0.914 & 217.3MB \\ \hline
\end{tabular}
\end{table}

The next step of the experiment is to merge detection with classification. For the classification subtask, we used the previously trained model SEN-EM-AB. Accuracy is computed as a number of correctly classified images, assuming that the classification is aggregated from all of the detected patches. The summary for the evaluation is in Tab.~\ref{tab:results-persondet}.  For comparison, we also show results for the end-to-end classifier and the classification made with the ground-truth detection coordinates. It can be noticed that the results are worse than those of the end-to-end classifier. This could mean that some of the features important for the end-to-end classifier lay outside the area occupied by humans, or the detector misses some of the important image parts. As the labels in the patch-based scenario are aggregated, recall for the sexual activity class improved compared to the end-to-end classifier.

\begin{table}
\caption{SE classification based on patches obtained from person detector. Accuracy is computed with respect to all images using aggregated classification.}
\label{tab:results-persondet}
\begin{tabular}{|p{0.16\textwidth}|p{0.36\textwidth}|p{0.11\textwidth}<{\raggedleft}|p{0.11\textwidth}<{\raggedleft}|p{0.11\textwidth}<{\raggedleft}|p{0.11\textwidth}<{\raggedleft}|}
\hline
\multirow{2}{*}{Scenario} & \multirow{2}{*}{Model} & \multirow{2}{0.11\textwidth}{Accuracy}  & \multirow{2}{0.11\textwidth}{F1-score}  &\multicolumn{2}{c|}{TPR} \\ \cline{5-6}
 & & & & Sexual activity & Sexual posing  \\ \hline
End-to-end classifier & SEN-EM-AB & 90.17\% & 0.912 & 72.55\% & 81.43\%  \\ \hline
\multirow{2}{0.16\textwidth}{Patch classification} & YoloX-L + SEN-EM-AB &  89.27\% & 0.909  & 74.81\% & 78.46\%  \\ \cline{2-6}
  & Ground-truth +SEN-EM-AB &  89.82\%   & 0.921
 & 79.05\%  & 80.85\%  \\ 
 \hline 	
\end{tabular}
\end{table}

\section{Private body parts detection}

As was noted before, the visibility of genitalia is not sufficient to distinguish between not-sexual and sexually-explicit images. In the group of neutral images, a large number of samples have visible intimate body parts without sexual context, like images of bathing children and nudist beach photos. However, if we want to find the particular difference between images that are described as sexual posing (which are illegal) and the ones described as erotic (which are legal), the presence of private body parts is the main factor here. Therefore, we decided to apply a body part detector as an additional method that can help classify potential SE content. In the proposed method, we look for the intimate body parts first, then inject this information into the classification system to determine if the image contains nudity. 

Our body part detector was trained on bounding boxes from the DN-A dataset. Similarly, as in the experiments with person detection, we've chosen the YoloX-L architecture. The proposed detector is trained to find one of the three selected body parts: male or female genitalia and anal area. Detection results are evaluated using previously mentioned metrics: average precision (AP) and average recall (AR) which are computed for IoU=0.5. In Tab.~\ref{tab:class-with-bodyparts}, results for detection are presented together with the accuracy of classification. In this task, classification is defined as a binary test of whether intimate body parts are visible or not. In Tab.~\ref{tab:class-with-bodyparts}, evaluation results are divided into three main SE categories as we intend to emphasize the difference between each group. The model achieves the worst accuracy for images belonging to the sexual activity class. The main reason could be the fact that in this group, there are pictures showing sexual intercourse where genitalia are not clearly visible. 
On the other hand, the model performs better on images belonging to the sexual posing class. The characteristics of images in that class are very consistent, and the genitals are usually distinctly visible. 
Model performance on images in neutral class is moderate. 
As a baseline, we show how many of the images in each group contain ground-truth annotations of body parts. It can be seen that more than 9\% of neutral images contain visible body parts. On the contrary, there are samples showing explicit sexual activity but without visible genitalia. There are more than 12\% of such samples in the category. That's why a private body part detector cannot be used as the only model for classification in our case.

\begin{figure}[!htb]
\centering
\includegraphics[width=\linewidth]{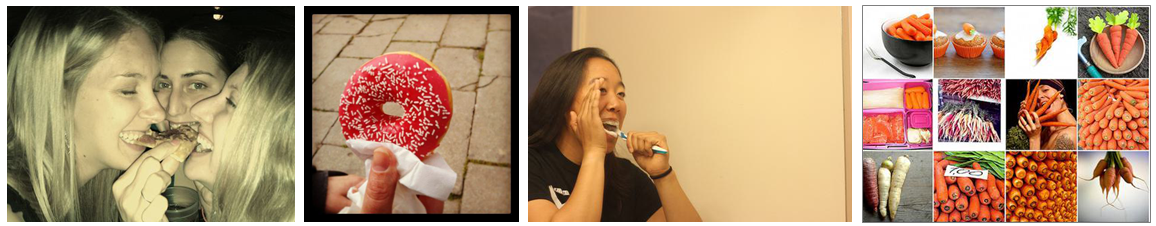}
\caption{Images from COCO dataset~\cite{coco2015} incorrectly classified as SE by the first version of a model trained only on DN-A (SEN-EM). A small number of neutral samples makes the model focus on images of people eating or keeping something near their mouths or images of close-up images of hands. False positives have been eliminated after adding a pretraining phase on extended datasets.}
\label{fig:cocofp}
\vspace*{1em}
\centering
\includegraphics[width=\linewidth]{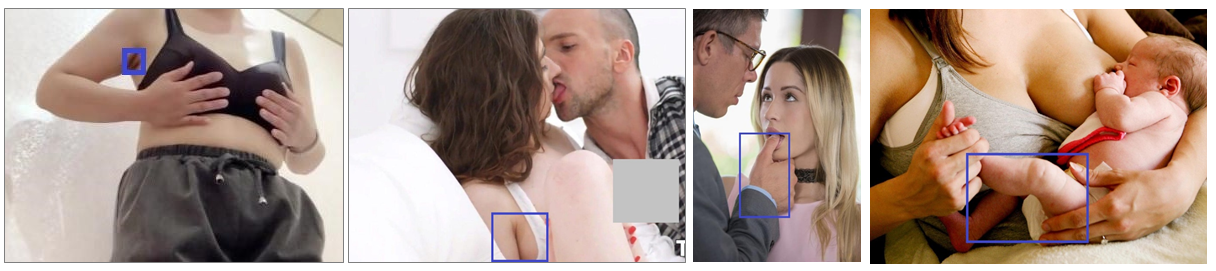}
\caption{False positive detection of neutral images from DN-A dataset. Body parts in dataset DN are usually small patches and can be challenging to detect. From left to right are bounding boxes of female genitalia, female genitalia, male genitalia, and male genitalia.}
\label{fig:fpgens}
\end{figure}

Detecting intimate body parts is a challenging task. They often occupy a relatively small area of the image. In our research, we noticed a problem with false detections, which are hard to reduce. Exemplary neutral images with incorrectly detected private body parts are presented in Fig.\ref{fig:fpgens}. In some photos, especially those with low quality, target body parts can be visually similar to other neutral body parts like armpits, lips, or fingers. Despite those problems, the proposed detector is a robust model. In the group of images tagged as "sexual posing," it achieves a high detection precision, and nudity in this group can be predicted with an accuracy equal to 91.22\%.

\begin{table}
\caption{Accuracy of body parts detection and classification of explicit content from information about the visibility of private body parts computed on dataset DN-A.}
\label{tab:class-with-bodyparts}
\begin{tabular}{|p{0.24\textwidth}|p{0.16\textwidth}<{\raggedleft}| p{0.18\textwidth}<{\raggedleft}|p{0.18\textwidth}<{\raggedleft}|p{0.18\textwidth}<{\raggedleft}|}
\hline
\multirow{2}{0.24\textwidth}{Label} & \multirow{2}{0.16\textwidth}{\% of samples with gt boxes} & Classification & \multicolumn{2}{c|}{Detection}\\ \cline{3-5}
 & & Accuracy & AP$^{IOU=0.5}$ &  AR\\ \hline
Neutral	& 9.28\% & 88.64\% & 0.2354 & 0.3978 \\
Sexual posing & 99.88\% & 91.22\% & 0.6951 & 0.7469 \\
Sexual activity	& 87.61\% & 76.72\% & 0.4921 & 0.5506 \\
All SE categories & 94.36\% & 84.70\% & 0.6103 & 0.6644 \\ \hline
All image categories & 63.03\% & 86.15\% & 0.5915 & 0.6530 \\
\hline
\end{tabular}
\end{table}

\section{Conclusions}

The article discusses approaches for classifying sexually explicit content with end-to-end neural networks and by applying detectors either for localizing people or intimate body parts in images.  Models were tested on real-world data collected by the team of experts, who manually inspect potentially explicit content as a part of their job. The aim of the work is to transfer their experience into the automated system and take off some work from human annotators.

The proposed classification system distinguishes 3 categories of images: showing sexual activity, sexual posing and neutral ones. Our schema was chosen based on data observation and leads to obtaining a more balanced evaluation dataset regarding the number of samples in each category. The following experiments have proven that the classification results also reflect the differences between categories. Sexual posing, for example, usually contains well-visible private body parts, so the methods of body parts detection can leverage the result for this particular class.

Despite the limited access to data, we discussed some results on neutral samples from the DN dataset or samples from external databases like COCO and Pornography-2K. During the analysis, 
we concluded that it is necessary to extend the training data, particularly with neutral samples, to improve the accuracy of results and reduce some of the false positives. We also experimented with collecting additional sets containing adult pornography. However, they seem to have different characteristics than the DN data, probably because of the varying definitions of sexual content. As adult pornography comes mostly from video footage, it contains a lot of frames in close-up and samples that are pornographic in the context of the sequence but not necessarily explicitly clear when analyzed alone. 

The highest accuracy was achieved using the end-to-end classifier, but we emphasize the importance of applying detectors in our projects to obtain more meaningful and explainable results. Focusing on human silhouette and body part detection is the right direction for further extending the proposed methods.
Overall, the study provides insights into different approaches for CSAM classification and highlights the challenges associated with the real-world CSAM detection system.

%

 \bibliographystyle{splncs04}
 \bibliography{prcnn}

\end{document}